\documentclass{article}

\usepackage[utf8]{inputenc}
\usepackage[T1]{fontenc}
\usepackage{CJKutf8}
\usepackage{newunicodechar}
\newunicodechar{，}{,}

\usepackage{longcat_style}

\usepackage{amsmath}
\usepackage{amsfonts}
\usepackage{amssymb}
\usepackage{bbm}
\usepackage{nicefrac}

\usepackage{xcolor}

\usepackage{array}
\usepackage{booktabs}
\usepackage{tabularx}
\usepackage{multirow}
\usepackage{makecell}
\usepackage{colortbl}
\usepackage{arydshln}

\usepackage{graphicx}
\usepackage{svg}
\usepackage{wrapfig}
\usepackage{subcaption}

\usepackage{enumitem}
\usepackage{microtype}
\usepackage{lipsum}
\usepackage{siunitx}
\usepackage{xspace}

\usepackage{url}
\usepackage{xurl}
\usepackage{natbib}
\usepackage{doi}

\usepackage{pifont}
\usepackage[most]{tcolorbox}

\usepackage{hyperref}
\usepackage{cleveref}

\hypersetup{
    colorlinks=true,
    linkcolor=blue,
    urlcolor=blue,
    citecolor=blue,
    linkbordercolor=blue,
    urlbordercolor=blue,
    citebordercolor=blue,
    pdfborderstyle={/S/U/W 1},
}

\setlist[itemize]{leftmargin=*}
\setlist[enumerate]{leftmargin=*}
\setlist[description]{leftmargin=*}

\definecolor{mygray}{gray}{.88}
\definecolor{mycyan}{cmyk}{.15,0,0,0}
\definecolor{mycyan2}{cmyk}{.85,0,0,0}
\definecolor{mygreen}{rgb}{0.19, 0.79, 0.02}
\definecolor{midnightgreen}{rgb}{0.0, 0.29, 0.33}
\definecolor{darkgreen}{RGB}{0,160,0}

\newcommand{\notcheckmark}{\textcolor{black}{\bcmark\kern-1.1ex\raisebox{.7ex}{\rotatebox[origin=c]{125}{--}}}\color{black}}
\newcommand{\bcmark}{\color{blue}{\ding{51}}}

\usepackage{tikz}
\usepackage{amsthm}
\usepackage{mdframed}

\usepackage[utf8]{inputenc} 
\usepackage{csquotes}

\usepackage{ulem}
\usepackage{caption}
\usepackage{multibib}
\usepackage{color}

\usepackage{float}
\usepackage{inconsolata}

\tcbuselibrary{listings,theorems}
\newtcolorbox{mybox}{colback=white!5!white,colframe=black!75!black, left=.05in, right=.05in}

\definecolor{bluex}{rgb}{0.27, 0.42, 0.81}
\definecolor{purplex}{HTML}{9564bf}
\definecolor{red3}{HTML}{C52A20}
\definecolor{red2}{HTML}{B36A6F}
\definecolor{red1}{HTML}{FFb5b5}
\definecolor{purple}{HTML}{B36A6F}
\definecolor{darkyellow}{HTML}{D5BA82}
\definecolor{blue1}{HTML}{508AB2}
\definecolor{blue2}{HTML}{C4E4E3}
\definecolor{green1}{HTML}{A1D0C7}
\definecolor{green2}{HTML}{BFF6BA}
\definecolor{green3}{HTML}{028100}
\definecolor{teal}{HTML}{508AB2}
\definecolor{purple1}{HTML}{8d3a94}

\newtcbtheorem[]{exmp}{Example}%
{colback=red2!5,colframe=red2!80,fonttitle=\bfseries, left=.08in, right=.08in,bottom=.05in, top=.05in}{exmp}

\title{\textsc{General365}: Benchmarking General Reasoning in Large Language Models Across Diverse and Challenging Tasks}

\author{
Junlin Liu\thanks{\, Work done during the internship at Meituan.}\,\,$^{\clubsuit}$, Shengnan An\thanks{\, Correspondence to: \texttt{\{anshengnan, caoxuezhi\}@meituan.com}.}\,\,$^{\diamondsuit}$, Shuang Zhou$^{\diamondsuit}$, Dan Ma$^{\diamondsuit}$, Shixiong Luo$^{\diamondsuit}$, Ying Xie$^{\diamondsuit}$, Yuan Zhang$^{\diamondsuit}$, \\
\textbf{Wenling Yuan$^{\diamondsuit}$, Yifan Zhou$^{\diamondsuit}$, Xiaoyu Li$^{\diamondsuit}$, Ziwen Wang$^{\diamondsuit}$, Xuezhi Cao$^{\dag \diamondsuit}$, Xunliang Cai$^{\diamondsuit}$}\\
\\
$^{\clubsuit}$University of Chinese Academy of Sciences\quad
$^{\diamondsuit}$Meituan\,\,\,
}

\renewcommand{\headeright}{\raisebox{-0.2\height}{\includegraphics[height=1.8em]{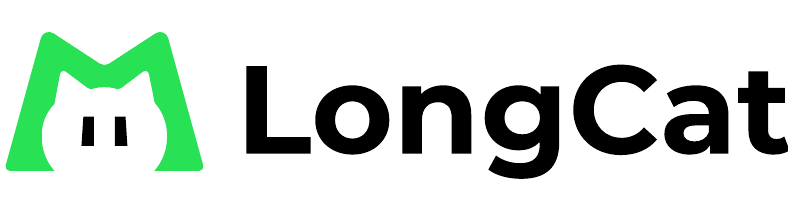}}}
\renewcommand{\shorttitle}{\fontsize{8.5}{11}\selectfont \textsc{General365}: Benchmarking General Reasoning in LLMs Across Diverse and Challenging Tasks}

\begin{document}
\maketitle

\begin{abstract}
Contemporary large language models (LLMs) have demonstrated remarkable reasoning capabilities, particularly in specialized domains like mathematics and physics.
However, their ability to generalize these reasoning skills to more general and broader contexts—often termed general reasoning—remains under-explored.
Unlike domain-specific reasoning, general reasoning relies less on expert knowledge but still presents formidable reasoning challenges, such as complex constraints, nested logical branches, and semantic interference.
To address this gap, we introduce \textbf{\textsc{General365}}, a benchmark specifically designed to assess general reasoning in LLMs. 
By restricting background knowledge to a K-12 level, \textsc{General365} explicitly decouples reasoning from specialized expertise.
The benchmark comprises 365 seed problems and 1,095 variant problems across eight categories, ensuring both high difficulty and diversity. 
Evaluations across 26 leading LLMs reveal that even the top-performing model achieves only 62.8\% accuracy, in stark contrast to the near-perfect performances of LLMs in math and physics benchmarks.
These results suggest that the reasoning abilities of current LLMs are heavily domain-dependent, leaving significant room for improvement in broader applications.
We envision \textsc{General365} as a catalyst for advancing LLM reasoning beyond domain-specific tasks toward robust, general-purpose real-world scenarios.
\end{abstract}

\begin{center}
\textbf{Code, Dataset, and Leaderboard:} { } \href{https://general365.github.io/}{\texttt{\textsc{General365}.github.io}}
\end{center}


\begin{figure}[ht]
    \centering
    \includegraphics[width=0.95\textwidth]{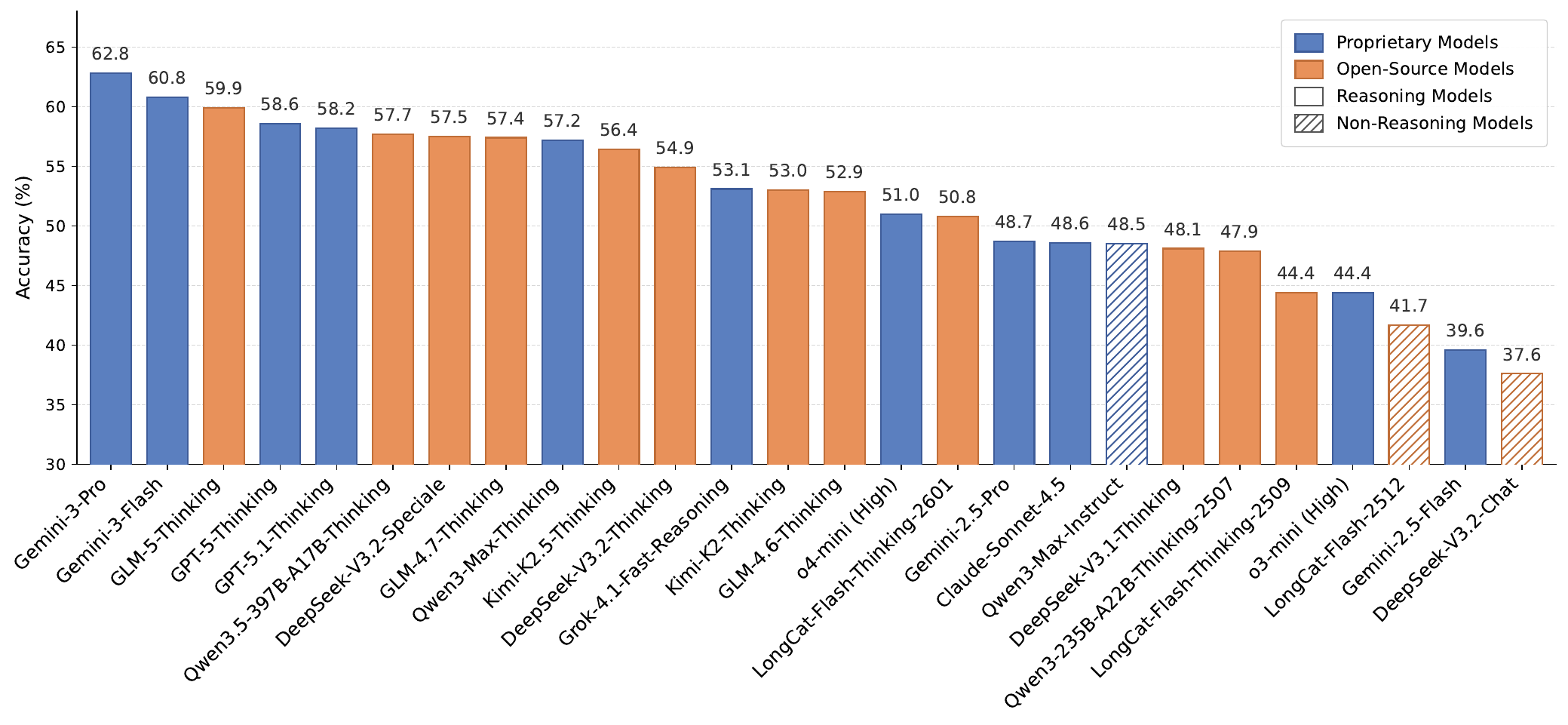} 
    \caption{Performance of various LLMs on \textsc{General365}. Gemini-3-Pro achieves state-of-the-art performance with 62.8\%, while the majority of models fail to reach the 60\% passing standard.}
    \label{fig:leaderboard}
\end{figure}

\section{Introduction}

Recently, Large Language Models (LLMs) have exhibited enhanced reasoning capacities across a diverse range of complex tasks~\citep{gemini3.1,gpt5.3,claude4.5,grok4.1,qwen3,deepseekr1,deepseekv3.2,longcatflash,glm5,seed2.0,hunyuan,kimik2.5,longcatflashthinking}, and how to evaluate their ability to solve such problems has become a key area of focus. 
Current evaluations of LLM reasoning capabilities focus on domains that rely on deep domain-specific knowledge such as mathematics, physics or programming~\citep{aime25, Physunibench, Swe-bench, phan2025humanity, gpqa, supergpqa}, where models have approached expert-level knowledge and reasoning performance. For instance, top-tier LLMs have achieved gold-medal-level scores on competitions like AIME, IMO, and IPhO ~\citep{gpt5.3, hubert2025olympiad, huang2026perfect}.


Despite the impressive performance of LLMs on reasoning competitions, these results do not necessarily translate to proficiency in real-world daily reasoning, which often relies on general logic rather than deep domain-specific knowledge. 
Consequently, there is a critical need to evaluate the general reasoning capabilities of LLMs within the foundational knowledge (e.g., K-12 level). 
While existing benchmarks~\citep{bigbench, bbh, bbeh, korbench} have made some attempts, the performance of current LLMs on these reveals two emerging challenges: (1) Less diversity: Some general benchmarks still have a limited variety of tasks, with highly similar construction templates for instances within tasks, with some tasks even exhibiting clear synthetic data features. This may lead to potential performance inflation due to reasoning shortcuts or hacks; (2) Insufficient difficulty: LLMs' performance on certain general benchmarks has reached a saturation point, making these benchmarks incapable of distinguishing between models in terms of their general reasoning abilities.


To address these limitations, we present \textsc{General365}, a benchmark specifically designed for evaluating general reasoning in LLMs.
The core features are as follows:
\begin{itemize}
    \item \textbf{High Diversity.}
    It contains 365 manually crafted, highly diverse seed problems, specifically designed to cover a wide range of reasoning challenges and avoid repetitive features or patterns. By altering surface semantics or constraints while preserving core reasoning skills, these seed problems were further expanded into 1,095 variants.
    \item \textbf{Challenging Boundaries.} \textsc{General365} covers 8 challenging categories, as detailed in Section ~\ref{sec:challenge}. Even state-of-the-art models barely achieve a "passing" level of performance on these challenging tasks.
    \item \textbf{Focus on Reasoning over Knowledge.} The knowledge required is strictly confined to the K-12 scope, ensuring the dataset measures a model's reasoning capabilities rather than knowledge retrieval.
    \item \textbf{Rigorous Quality Control.} All instances have undergone manual review to ensure the highest standards of quality.
    \item \textbf{Accurate Scoring.} We implemented a hybrid scoring algorithm combining rule-based and model-based approaches, achieving a manually verified scoring accuracy of 99.6\%.
\end{itemize}


Experimental results across various LLMs demonstrate that contemporary LLMs still have considerable room for improvement in general reasoning tasks within real-world, everyday domains presented by \textsc{General365}.
Among the 26 evaluated models, the top-performing model, Gemini-3-Pro, achieved an accuracy of only 62.8\% on \textsc{General365}, with most models failing to meet the passing standard.
Moreover, we further analyze the reasoning efficiency of various LLMs on the \textsc{General365} benchmark. 
Notably, Gemini-3-Pro achieves state-of-the-art performance while maintaining a remarkably concise average output of approximately 15k tokens—a significant reduction compared to the 25k tokens typically required by other top-tier frontier models. This empirical evidence highlights its superior reasoning density, demonstrating an ability to resolve complex tasks with high reasoning efficiency. 
To further delineate the performance bottlenecks of current LLMs, we conduct a fine-grained evaluation across eight distinct challenge categories. 
Our analysis reveals that while the models exhibit relatively balanced performance across most challenge categories, \textit{"Semantic Interference"} and \textit{"Optimal Strategy"} emerge as the primary challenges for current LLMs. 
These findings delineate the current general reasoning boundaries of LLMs.

Crucially, the diversity of \textsc{General365} was rigorously verified on seed problems through a dual-faceted approach, ensuring that the benchmark spans a sufficiently broad semantic and logical spectrum. (1) Qualitatively, our query visualization via embedding projection reveals that \textsc{General365} exhibits a broad and relatively uniform distribution across the semantic space, while BBH and BBEH demonstrate significant local clustering. This spatial dispersion indicates that \textsc{General365} not only covers a diverse range of reasoning scenarios but also maintains sufficient discriminative granularity between individual tasks, effectively preventing semantic redundancy.
(2) Quantitatively, we evaluated the logical redundancy among problem instances by leveraging LLMs to score the similarity of reasoning paths between cases. The results demonstrate that \textsc{General365} exhibits a significantly lower logical overlap score compared to BBH and BBEH. This marked reduction in inter-case similarity underscores the superior diversity of \textsc{General365} in terms of reasoning primitives and problem-solving strategies.

The \textsc{General365} benchmark is publicly available at \href{https://general365.github.io/}{\texttt{\textsc{General365}.github.io}}.
By providing this diverse and challenging benchmark, we aim to facilitate further application on robust, general-purpose real-world scenarios.

\section{\textsc{General365}}\label{sec:general365}
\begin{figure}[t]
    \centering
    \includegraphics[width=0.87\textwidth]{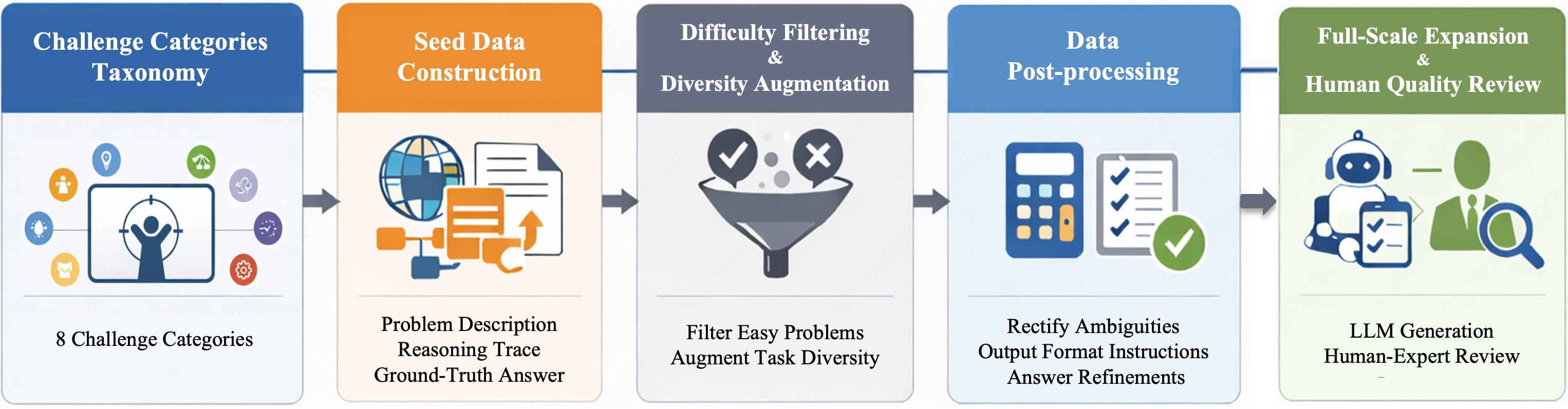} 
    \caption{The construction pipeline of \textsc{General365}.}
    \label{fig:pipeline}
\end{figure}

In this section, we first introduce the challenge categories taxonomy of \textsc{General365} (Section~\ref{sec:challenge}), followed by a detailed description of the data construction pipeline (Section~\ref{sec:data_construct}).
We then present the dataset statistics and elaborate on the grading methodology, including the evaluation accuracy specifically designed for \textsc{General365} (Section~\ref{sec:grading}). 
Figure~\ref{fig:pipeline} illustrates the construction and grading pipeline of \textsc{General365}.


\subsection{Challenge Categories Taxonomy}\label{sec:challenge}
To ensure high-level structural diversity and pinpoint the performance bottlenecks of current LLMs in general reasoning, we establish a conceptual taxonomy for the challenge covered by \textsc{General365}. 
Specifically, by synthesizing existing reasoning benchmarks and analyzing real-world inferential requirements, we have codified eight distinct challenge categories. 
The definitions and detailed descriptions of these categories are summarized below, with representative examples detailed in Appendix ~\ref{appendix:examples}.

\begin{itemize}
    \item \textbf{Complex Constraints.} These tasks involve a web of interconnected logical predicates that must be satisfied simultaneously. To resolve them, LLMs must maintain state-tracking across multi-step reasoning chains, ensuring that local inferences remain globally consistent.
    \item \textbf{Branching \& Enumeration.} Success in this category requires LLMs to perform a systematic exploration of complex solution spaces. Models must demonstrate branching case analysis and exhaustive reasoning capabilities to ensure that no marginal scenarios or latent boundary conditions are overlooked.
    \item \textbf{Spatial \& Temporal Reasoning.} This category evaluates LLMs' proficiency in spatial manipulation or temporal progression, such as geometric arrangements and causal sequencing. By shifting the focus to dynamic reasoning, this challenge requires LLMs to model how spatial properties or temporal states evolve under specific sequences of operations.
    \item \textbf{Recursive \& Backtracking.} Designed to evaluate non-linear problem-solving abilities, the core of this category is self-correction. These tasks require LLMs to formulate hypotheses and engage in an iterative process of trial, verification, and backtracking. 
    \item \textbf{Semantic Interference.} These problems introduce cognitive traps by setting rules that defy common-sense intuition or by embedding misleading information. They evaluate the model’s ability to suppress pre-trained factual biases and adhere strictly to a set of novel axioms during the reasoning process.
    \item \textbf{Implicit Information Reasoning.} Rather than providing all premises explicitly, these tasks require clue discovery. LLMs must bridge disjointed observations (e.g., symbolic patterns or latent mathematical regularities) to infer the underlying logical framework.
    \item \textbf{Optimal Strategy.} Moving beyond simple correctness, this category focuses on decision-making under constraints. LLMs are tasked with evaluating multiple viable action sequences to determine the most efficient path, testing their capacity for utility-based optimization and strategic planning.
    \item \textbf{Probability \& Uncertainty.} These tasks are designed to evaluate the ability of LLMs to reason within stochastic environments characterized by incomplete information. By requiring LLMs to estimate the likelihood of complex propositions or assess their logical possibility, this category shifts the focus from binary to probabilistic inference and uncertainty quantification.
\end{itemize}

\subsection{Dataset Construction Pipeline}\label{sec:data_construct}
Guided by the challenge categories taxonomy introduced above, we developed a multi-stage construction pipeline to ensure the diversity, complexity and quality of \textsc{General365}. The workflow consisted of four primary phases:


\paragraph{Seed Data Construction.} 
We crowdsource problems from diverse, real-world domains to ensure the inherent diversity of the seed data. 
Additionally, to guarantee the high quality and integrity of \textsc{General365}, each problem must satisfy the following criteria: (1) Manually verified to align with at least one of the eight defined challenge categories. (2) Must be original and non-replicable via standard web searches. (3) Seed instance is structured as a triplet, comprising the problem description, detailed reasoning trace, and the ground-truth answer. (4) A final round of manual inspection is conducted to cross-verify the consistency and correctness of the problem, gold answer and reasoning path.


\paragraph{Difficulty Filtering \& Diversity Augmentation.}
Following the collection of seed data, we conducted a rigorous filtering process to eliminate easy or similar instances. Specifically, we filtered out problems that were easily solvable by top-tier models or demonstrated high semantic similarity to existing benchmarks. This ensures the remaining seed set remains non-trivial and provides a meaningful challenge for advanced reasoning models.
Furthermore, we augmented the dataset by manually expanding categories with limited samples.


\paragraph{Data Post-processing.} 
To ensure that LLMs accurately interpret task requirements and generate standardized outputs for reliable evaluation, we implemented a post-processing phase prior to the final full-scale expansion. This refinement process focuses on the following three dimensions: (1) By analyzing the interactions between LLMs and seed data, we identified and rectified potential linguistic ambiguities, ensuring that the task premises are clear and unmistakable.
 (2) We appended output format instructions to each problem description (e.g., "Select one or more appropriate options as your final answer based on the question above."), guiding LLMs to adhere to a structured output protocol. 
 (3) Wherever feasible, we restructured answers into numerical or canonical formats, facilitating robust, rule-based extraction and grading.


\paragraph{Full-Scale Expansion \& Human Quality Review.} 
To scale the dataset while maintaining high diversity, we implemented an LLM-based expansion strategy with human-in-the-loop. 
We initially leveraged LLMs to generate 10 candidate expansions for each seed problem. Subsequently, human-experts performed a stringent quality review, discarding incorrect or illogical entries. 
For seed problems with fewer than three valid expansions, manual intervention was employed to supplement the data. 
Finally, we performed a comprehensive difficulty estimation and diversity audit to select the optimal candidates, resulting in a diverse and challenging benchmark: \textsc{General365}.

\begin{exmp}{Example of General Reasoning Problem}{numerical_answer}
{}\textbf{Question:} Strangers A, B, C, D, and E line up from youngest on the left to oldest on the right. Their clothing colors and shoe colors all differ, and they come from five different regions. \par
Known facts: \par
~~~~ 1. A is from Morocco. \par
~~~~ 2. D is five years older than B. \par
~~~~ 3. E is older than A. \par
~~~~ 4. C stands next to D. \par
~~~~ 5. A stands next to B. \par
~~~~ 6. The person in teal shoes is not adjacent to the person from Vanuatu. \par
~~~~ 7. One twelve-year-old wears yellow shoes. \par
~~~~ 8. The person in orange shoes wears white clothing. \par
~~~~ 9. The person in blue clothing is from Chile. \par
~~~~ 10. The youngest person wears red shoes. \par
~~~~ 11. Counting from the right, the fourth person comes from South Africa. \par
~~~~ 12. E wears yellow clothing. \par
~~~~ 13. The person in green shoes does not wear multicolored clothing. \par
~~~~ 14. Two people are twelve years old, ordered by birth month. \par
~~~~ 15. One adult is thirty-five years old, and that age is sixteen less than the combined ages of the other four. \par
If you multiply every possible age C might have, what number do you obtain? \par
\textbf{Answer:} \boxed{420}
\end{exmp}


\begin{figure}[t]
  \centering
  \begin{subfigure}[b]{0.58\textwidth}
    \centering
    \includegraphics[width=0.9\textwidth]{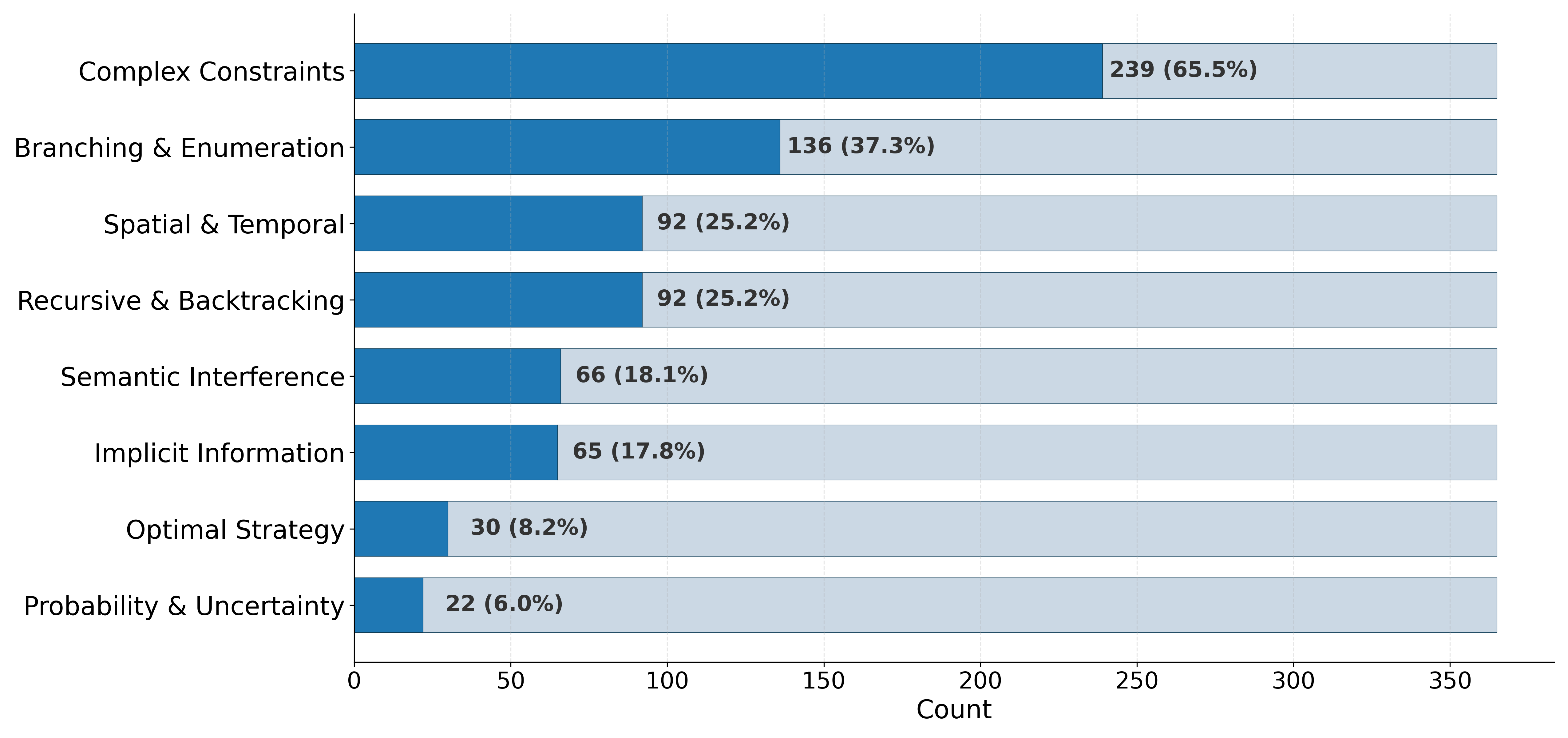}
    \caption{Distribution of challenge categories.}
    \label{fig:challenge}
  \end{subfigure}
  \hfill
  \begin{subfigure}[b]{0.38\textwidth}
    \centering
    \includegraphics[width=0.99\textwidth]{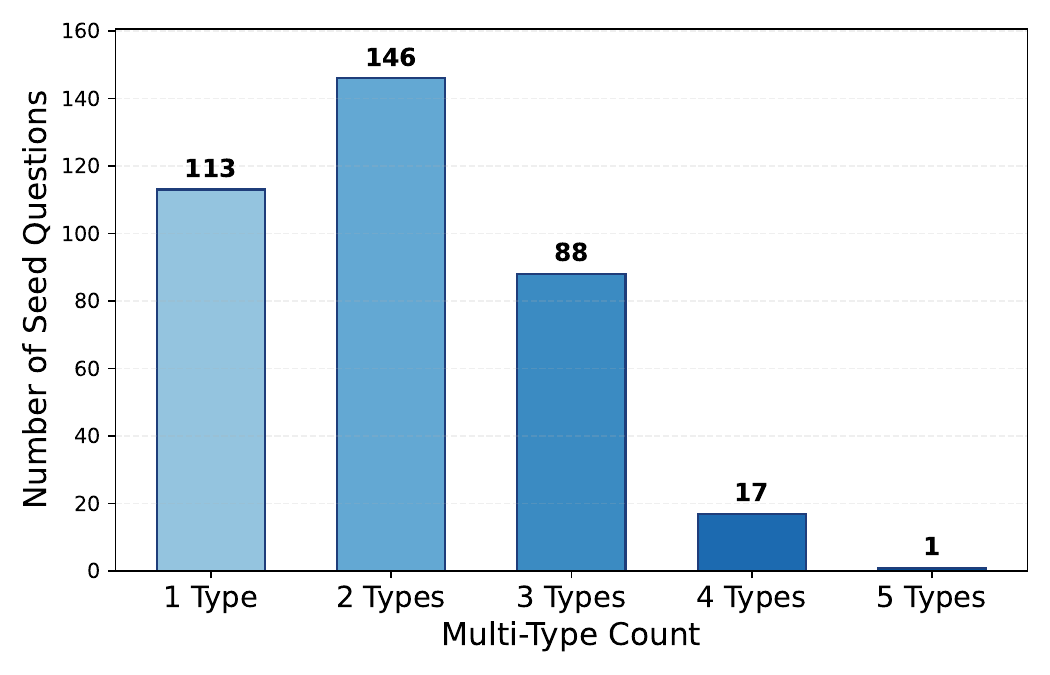}
    \caption{Distribution of multi-label questions.}
    \label{fig:challenge_label}
  \end{subfigure}
  \caption{Statistical Overview of \textsc{General365}. (a) Distribution of challenge categories, highlighting the diversity of problems within the benchmark. (b) Distribution of multi-label problems, showcasing the complexity in task assignments.}
  \label{fig:two_images_sub}
\end{figure}

\subsection{Dataset Statistics}\label{sec:statistic}
Based on the challenge categories taxonomy above, each seed problem in \textsc{General365} is annotated with one or more challenge labels. 
Figure ~\ref{fig:challenge} illustrates the overall distribution of challenge categories within \textsc{General365}. As shown, \textit{"Complex Constraints"} emerges as the most predominant challenge, while \textit{"Probability \& Uncertainty"} accounts for the smallest proportion of the dataset. 
Despite being the least represented category, \textit{"Probability \& Uncertainty"} still contains over 20 seed problems, ensuring a requisite level of intra-category diversity. This sample size is strategically maintained to be at least comparable to established benchmarks such as BBH and BBEH~\citep{bbh,bbeh}.
Furthermore, Figure~\ref{fig:challenge_label} presents the distribution of multi-label instances. Notably, nearly 70\% of the problems are annotated with two or more challenge tags, further underscoring that \textsc{General365} is characterized by high-difficulty, composite challenges rather than isolated reasoning tasks.

\subsection{Grading Method}\label{sec:grading}
Existing evaluation paradigms for LLMs primarily fall into two categories: rule-based and model-based grading. Rule-based approaches offer substantial computational efficiency and accuracy, provided the model responses adhere to parseable structures. 
Conversely, model-based grading provides superior flexibility, enabling the nuanced assessment of complex, open-ended responses. To achieve an optimal balance between evaluation accuracy and computational efficiency, we adopt a hybrid grading framework that dynamically selects the grading method based on the specific answer type of each problem. 
Specifically, problems in \textsc{General365} are divided into three main answer types: numerical answers, select answers and text answers.

For problems requiring numerical answers ($238/365$), we mandate the use of the LaTeX command as \texttt{\textbackslash boxed\{<answer>\}} to encapsulate the final answer. 
Additionally, for problems involving numerical approximations, we manually annotated the required precision threshold, specifying the minimum number of decimal places for a response to be considered correct. 
We finally utilize the tools provided by \texttt{math-verify}\footnote{\href{https://github.com/huggingface/Math-Verify}{https://github.com/huggingface/Math-Verify}.} to parse these answers and verify the equivalence with the ground truth. For problems requiring select or text answers ($46/365$ and $81/365$), we use model-based grading with GPT-4.1 serving as the grading model. The prompt templates used for the model-based grading method are shown in Appendix ~\ref{sec:prompts}.

\paragraph{Grading accuracy.}
To verify the reliability of our hybrid grading method, we conducted a rigorous manual quality review prior to the full-scale evaluation. 
We reviewed a total of 1,460 grading instances based on the responses of four representative frontier models (e.g., o4-mini, Gemini-2.5-Pro, DeepSeek-V3.1-Thinking, and LongCat-Flash-Thinking) on the seed problems. 
Our manual verification revealed a grading accuracy of 99.6\%, demonstrating that the proposed framework provides a highly precise and robust measure of model performance on \textsc{General365}.

\section{Experiments and Analysis}
In this section, we present the experimental results of several top-tier LLMs on \textsc{General365}. 
First, we describe the experimental setup (Section~\ref{sec:setup}), followed by a comprehensive discussion of the main results and further analysis (Section~\ref{sec:exp_results}).
Subsequently, we validate the high diversity and challenge of \textsc{General365} (Section~\ref{sec:validation}).

\subsection{Experimental Setup}\label{sec:setup}

\paragraph{Models.}
To conduct a granular and comprehensive evaluation of \textsc{General365}, we curate a diverse ensemble of premier LLMs from leading providers, including OpenAI, Gemini, Anthropic, DeepSeek, Qwen, GLM, Moonshot, and LongCat.
This selection encompasses a representative spectrum of both proprietary and open-source models. 
Additionally, beyond reasoning-native architectures specifically engineered for complex reasoning tasks, we intentionally incorporate several high-performance non-reasoning models like Qwen3-Max-Instruct, LongCat-Flash.
By benchmarking these models, we aim to explore the boundaries of native base model reasoning abilities without optimizations for "Long-Term Thinking" or "Test-Time Scaling."

\paragraph{Inference settings}
To comprehensively explore the reasoning trajectories when facing the diverse tasks of \textsc{General365}, we set the temperature to \texttt{T} = 1.0 for reasoning models and \texttt{T} = 0.7 for non-reasoning models. 
For all experiments, we fixed \texttt{top-p} = 1.0, left \texttt{top-k} unconstrained, and set the maximum output length to each model’s capacity. 
Notably, for models that support adjustable reasoning effort or budget (e.g., Test-Time Scaling), we defaulted to the highest available setting to elicit their peak reasoning performance. 


\subsection{Main Results}~\label{sec:exp_results}

\begin{table*}[t]
\centering
\renewcommand{\arraystretch}{1.1}
\caption{Accuracy of \textsc{General365} across various challenge categories for different models.}
\label{tab:challenge_results}
\resizebox{\textwidth}{!}{
    \begin{tabular}{>{\bfseries}cccccccccc}
\toprule
\textbf{Model} & \textbf{\begin{tabular}[c]{@{}c@{}}Complex \\Constraints\end{tabular}} & \textbf{\begin{tabular}[c]{@{}c@{}}Branching \\\& Enumeration\end{tabular}} & \textbf{\begin{tabular}[c]{@{}c@{}}Spatial \\\& Temporal\end{tabular}} & \textbf{\begin{tabular}[c]{@{}c@{}}Recursive \\\& Backtracking\end{tabular}} & \textbf{\begin{tabular}[c]{@{}c@{}}Semantic \\Interference\end{tabular}} & \textbf{\begin{tabular}[c]{@{}c@{}}Implicit \\Information\end{tabular}} & \textbf{\begin{tabular}[c]{@{}c@{}}Optimal \\Strategy\end{tabular}} & \textbf{\begin{tabular}[c]{@{}c@{}}Probability \\\& Uncertainty\end{tabular}} & \textbf{Overall} \\ 
\midrule
\multicolumn{10}{c}{\textit{\textcolor{orange}{\textbf{Reasoning Models}}}} \\
\midrule
Gemini-3-Pro & \textbf{65.3\%} & 64.3\% & \textbf{57.6\%} & 66.8\% & 55.7\% & \textbf{68.8\%} & 50.8\% & 54.5\% & \textbf{62.8\%} \\
Gemini-3-Flash & 62.9\% & 64.5\% & 56.5\% & \textbf{69.3\%} & 55.7\% & 62.7\% & 48.3\% & \textbf{60.2\%} & 60.8\% \\
GLM-5-Thinking & 63.5\% & \textbf{69.1\%} & 52.9\% & 68.2\% & 49.2\% & 55.4\% & \textbf{51.7\%} & 56.8\% & 59.9\% \\
GPT-5-Thinking & 61.7\% & 65.4\% & 48.6\% & 65.2\% & \textbf{57.2\%} & 58.5\% & 47.5\% & 59.1\% & 58.6\% \\
GPT-5.1-Thinking & 61.1\% & 62.9\% & 48.6\% & 64.4\% & 53.4\% & 61.5\% & 47.5\% & 47.7\% & 58.2\% \\
Qwen3.5-397B-A17B-Thinking & 60.1\% & 62.1\% & 50.0\% & 67.9\% & 50.8\% & 59.2\% & 43.3\% & 59.1\% & 57.7\% \\
DeepSeek-V3.2-Speciale & 60.4\% & 63.6\% & 52.2\% & 65.8\% & 51.5\% & 51.9\% & 50.0\% & 55.7\% & 57.5\% \\
GLM-4.7-Thinking & 59.1\% & 63.8\% & 49.7\% & 68.8\% & 49.6\% & 57.3\% & 49.2\% & 58.0\% & 57.4\% \\
Qwen3-Max-Thinking & 59.4\% & 65.3\% & 51.4\% & 66.8\% & 46.2\% & 60.0\% & 48.3\% & 53.4\% & 57.2\% \\
Kimi-K2.5-Thinking & 60.0\% & 64.3\% & 46.5\% & 65.2\% & 47.0\% & 54.6\% & 48.3\% & 54.5\% & 56.4\% \\
DeepSeek-V3.2-Thinking & 57.1\% & 62.5\% & 45.1\% & 64.9\% & 49.2\% & 52.3\% & 48.3\% & 53.4\% & 54.9\% \\
Grok-4.1-Fast-Reasoning & 56.6\% & 59.9\% & 43.8\% & 62.5\% & 50.8\% & 50.0\% & 39.2\% & 46.6\% & 53.1\% \\
Kimi-K2-Thinking & 54.9\% & 61.9\% & 44.6\% & 62.5\% & 42.8\% & 53.1\% & 46.7\% & 52.3\% & 53.0\% \\
GLM-4.6-Thinking & 57.2\% & 59.9\% & 40.8\% & 64.1\% & 48.5\% & 55.4\% & 38.3\% & 43.2\% & 52.9\% \\
o4-mini & 54.2\% & 57.7\% & 38.6\% & 62.8\% & 45.5\% & 53.5\% & 40.8\% & 47.7\% & 51.0\% \\
LongCat-Flash-Thinking-2601 & 54.1\% & 61.4\% & 38.3\% & 59.5\% & 45.1\% & 46.2\% & 39.2\% & 51.1\% & 50.8\% \\
Gemini-2.5-Pro & 50.9\% & 56.1\% & 35.9\% & 57.6\% & 48.1\% & 48.1\% & 34.2\% & 44.3\% & 48.7\% \\
Claude-Sonnet-4.5 & 51.9\% & 55.7\% & 37.2\% & 58.2\% & 47.3\% & 47.7\% & 37.5\% & 35.2\% & 48.6\% \\
DeepSeek-V3.1-Thinking & 51.3\% & 56.8\% & 38.3\% & 58.4\% & 43.6\% & 42.3\% & 45.8\% & 50.0\% & 48.1\% \\
Qwen3-235B-Thinking-2507 & 50.8\% & 55.1\% & 32.3\% & 60.9\% & 49.2\% & 46.5\% & 41.7\% & 45.5\% & 47.9\% \\
LongCat-Flash-Thinking-2509 & 48.5\% & 54.2\% & 30.7\% & 54.3\% & 44.7\% & 37.3\% & 38.3\% & 39.8\% & 44.4\% \\
o3-mini & 47.4\% & 51.1\% & 32.9\% & 54.6\% & 38.6\% & 47.3\% & 28.3\% & 30.7\% & 44.4\% \\
Gemini-2.5-Flash & 42.8\% & 46.5\% & 28.3\% & 47.8\% & 35.6\% & 33.8\% & 35.8\% & 34.1\% & 39.6\% \\
\midrule
\multicolumn{10}{c}{\textit{\textcolor{blue}{\textbf{Chat Models}}}} \\
\midrule
Qwen3-Max-Instruct & 52.0\% & 56.3\% & 33.2\% & 58.2\% & 47.7\% & 48.1\% & 40.8\% & 47.7\% & 48.5\% \\
LongCat-Flash-2512 & 45.3\% & 51.8\% & 26.6\% & 57.6\% & 40.9\% & 34.6\% & 27.5\% & 44.3\% & 41.7\% \\
DeepSeek-V3.2-Chat & 40.7\% & 46.9\% & 26.1\% & 50.3\% & 37.1\% & 29.6\% & 22.5\% & 35.2\% & 37.6\% \\
\bottomrule
\end{tabular}
}
\end{table*}

\paragraph{Leaderboard.}
Figure ~\ref{fig:leaderboard} presents the performance of 26 leading LLMs on \textsc{General365}, categorized by their proprietary vs. open-source status and reasoning vs. non-reasoning architectures. 
The experimental results reveal that contemporary LLMs still face significant challenges in addressing the complex, multi-step reasoning tasks within real-world, everyday domains covered by the \textsc{General365} benchmark. 
Even the state-of-the-art model, Gemini-3-Pro, achieves an accuracy of only 62.8\%, while the vast majority of evaluated models fail to reach a basic passing threshold. 
Following closely, Gemini-3-Flash and GPT-5-Thinking occupy the top tier with scores of 60.8\% and 58.6\%, respectively. 
In the open-source landscape, reasoning models have demonstrated remarkable competitiveness, narrowing the performance gap with the proprietary models. GLM-5-Thinking leads the open-source category with a score of 59.9\%, the fact that the best-performing open-source model is within a 3\% margin of the top proprietary model indicates that recent advancements in the open-source community are rapidly approaching the frontiers of commercial reasoning capabilities. 
Notably, some non-reasoning models, such as Qwen3-Max-Instruct and LongCat-Flash, outperform several dedicated reasoning models (e.g., Gemini-2.5-Flash), demonstrating their significant potential in tackling complex reasoning tasks.

\begin{figure}[t]
    \centering
    \includegraphics[width=0.75\textwidth]{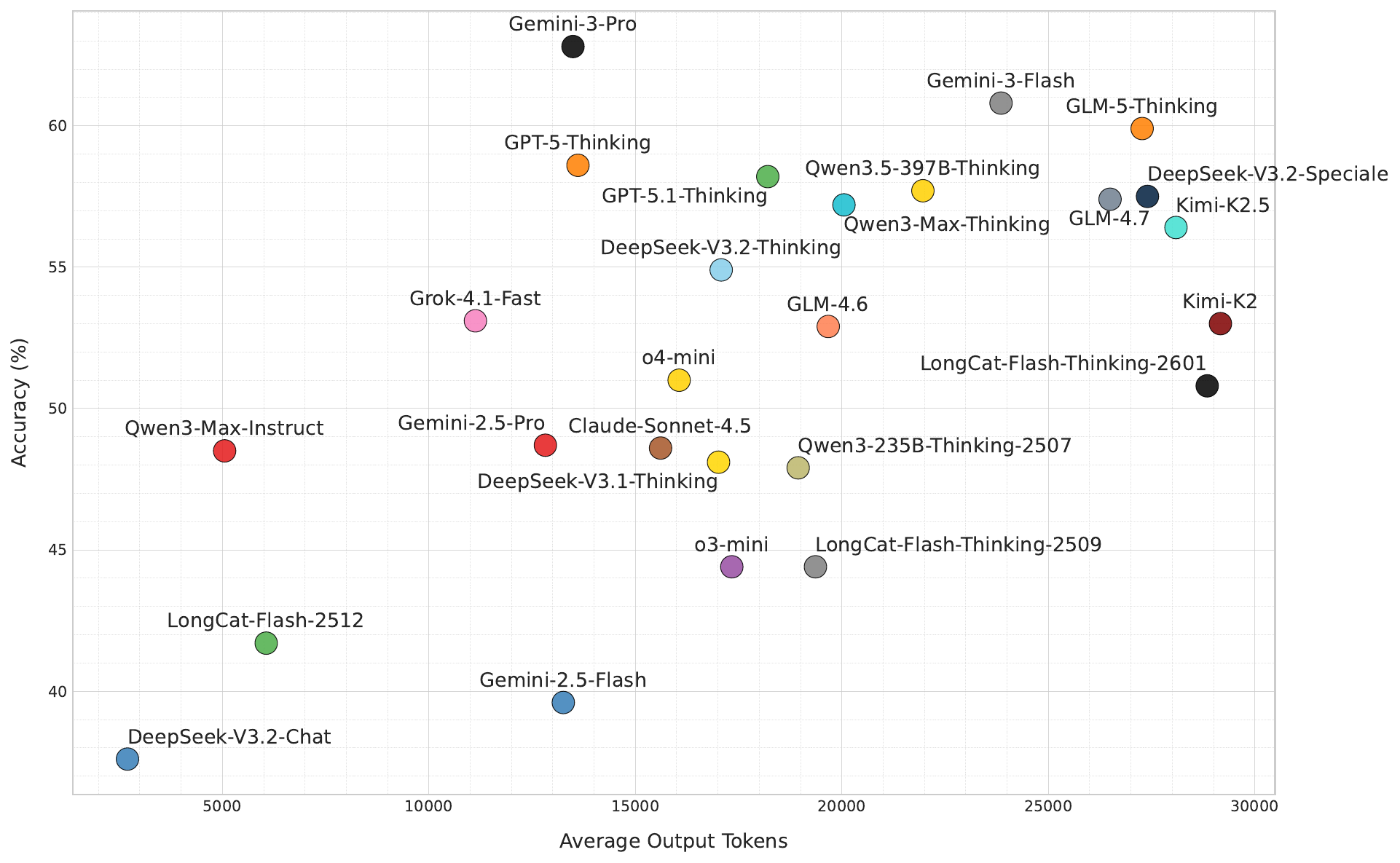} 
    \caption{The relationship between accuracy and average output tokens for various LLMs, highlighting the reasoning efficiency on the \textsc{General365} benchmark. Gemini-3-Pro exhibits superior reasoning efficiency compared to other frontier models.}
    \label{fig:token_efficient}
\end{figure}

\begin{wrapfigure}{r}{0.5\textwidth}
    \centering
    \includegraphics[width=0.53\textwidth]{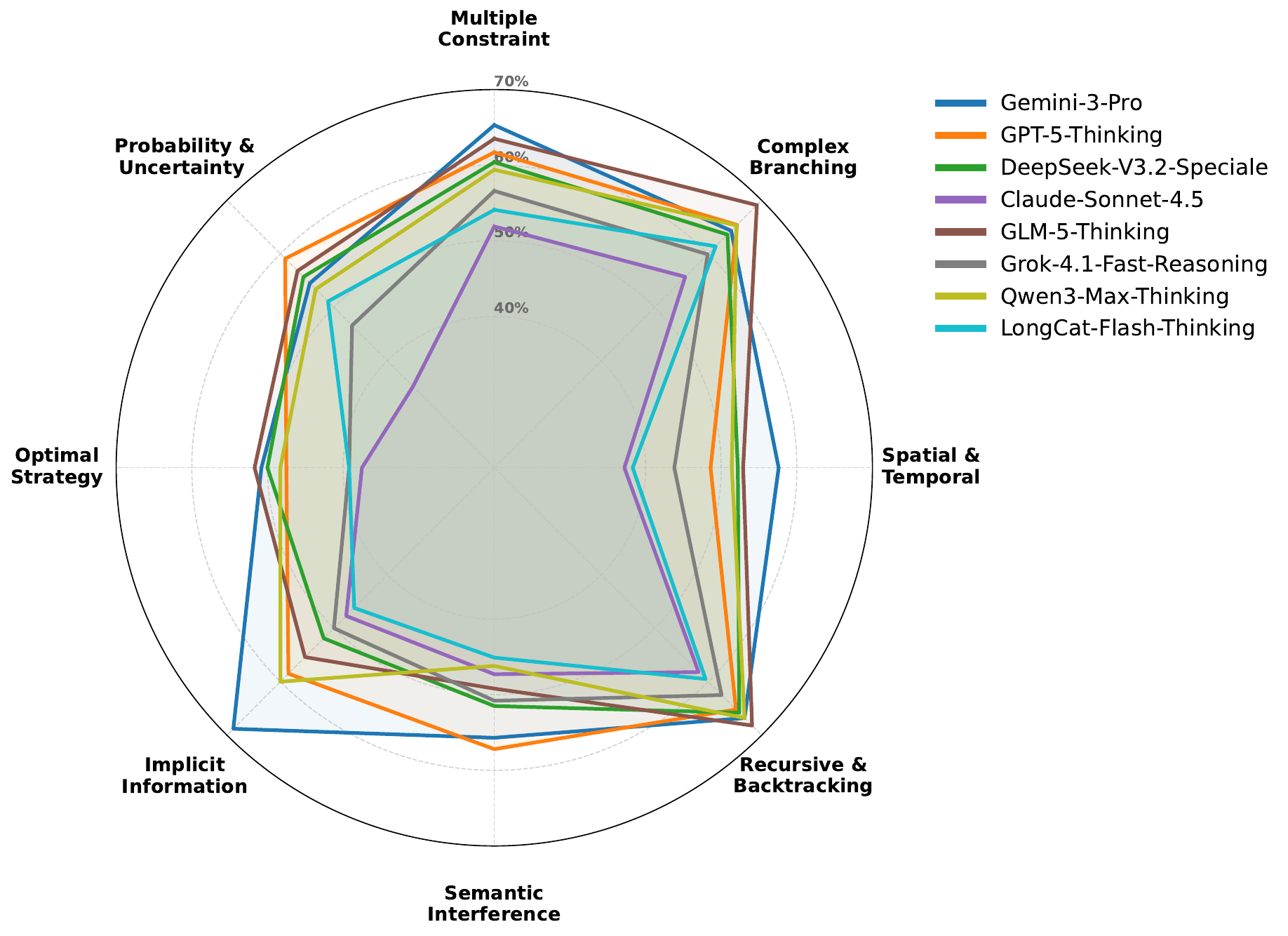} 
    \caption{Radar chart of various LLMs series across eight challenge categories.}
    \label{fig:radar}
\end{wrapfigure}

\paragraph{"Semantic Interference" and "Optimal Strategy" emerge as the primary performance bottlenecks for current LLMs.} 
To gain a granular understanding of the specific cognitive boundaries of current LLMs, we conduct a fine-grained analysis of model performance across eight distinct challenge categories provided by section ~\ref{sec:challenge}. 
Overall, the experimental results demonstrate a relatively balanced distribution of difficulty across the categories, with no single dimension appearing exceptionally trivial or insurmountable. As shown in Table \ref{tab:challenge_results}, models that achieve SOTA performance in specific challenge categories also tend to rank high in overall accuracy, suggesting that general reasoning proficiency is a composite of these foundational abilities. 
Despite this balance, \textit{"Semantic Interference"} and \textit{"Optimal Strategy"} emerge as the most significant hurdles. Performance in these two categories is consistently approximately 10\% lower than the models' overall average accuracy. This indicates that contemporary LLMs struggle with long-term strategic planning and are notably susceptible to distracting information or cognitive traps, which often divert the model from the core logical constraints. 
Moreover, in categories like \textit{"Implicit Information"}, the gap between top Reasoning Models (e.g., Gemini-3-Pro at 68.8\%) and Chat Models (e.g., DeepSeek-V3.2-Chat at 29.6\%) is at its widest. This confirms that the reasoning models excel at decoding subtle, non-explicit signals into structured logical frameworks that chat models often fail to capture.

\paragraph{Comparison of reasoning efficiency.}
Figure \ref{fig:token_efficient} illustrates the relationship between performance and average output token length on the \textsc{General365} benchmark. Our analysis reveals several key insights into the reasoning efficiency of current LLMs. (1) Overall, a positive correlation is observed between model performance and output token length. Top-tier frontier models generally utilize a high volume of tokens to navigate the complex reasoning paths required by \textsc{General365}. For instance, several leading reasoning models, such as DeepSeek-V3.2-Speciale and GLM-5-Thinking, require an average of approximately 25k to 30k tokens to achieve high accuracy. This trend suggests that current architectures often rely on extended inference-time computation to resolve multi-step reasoning challenges. (2) A notable exception to this trend is Gemini-3-Pro, which achieves SOTA performance (62.8\% accuracy) while maintaining a remarkably concise average output of approximately 14k tokens. 
This empirical evidence highlights its superior reasoning efficiency, enabling it to resolve complex tasks with high accuracy using fewer computational resources. (3) Furthermore, within specific model series, we observe clear trajectories of improved reasoning efficiency over time. For example, Gemini-3-Pro demonstrates a significant performance leap over Gemini-2.5-Pro while maintaining a comparable and efficient token expenditure.

\subsection{Validation of \textsc{General365} Features}\label{sec:validation}

\subsubsection{Validation of Diversity}
\begin{figure}[t]
  \centering
  \begin{subfigure}[b]{0.33\textwidth}
    \centering
    \includegraphics[width=0.99\textwidth]{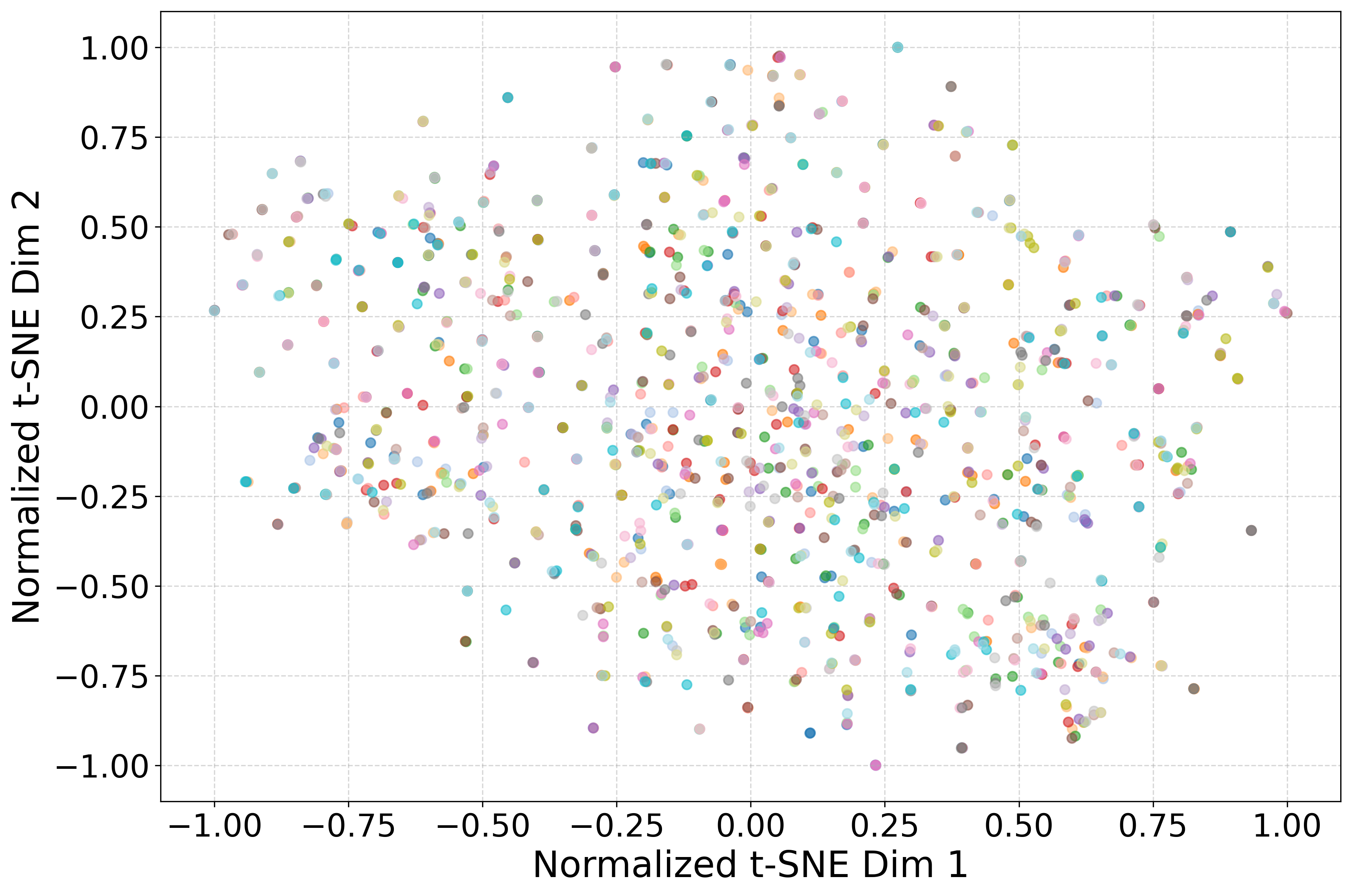}
    \caption{\textsc{General365} embeddings.}
    \label{fig:general365_embedding}
  \end{subfigure}
  \hfill
  \begin{subfigure}[b]{0.33\textwidth}
    \centering
    \includegraphics[width=0.99\textwidth]{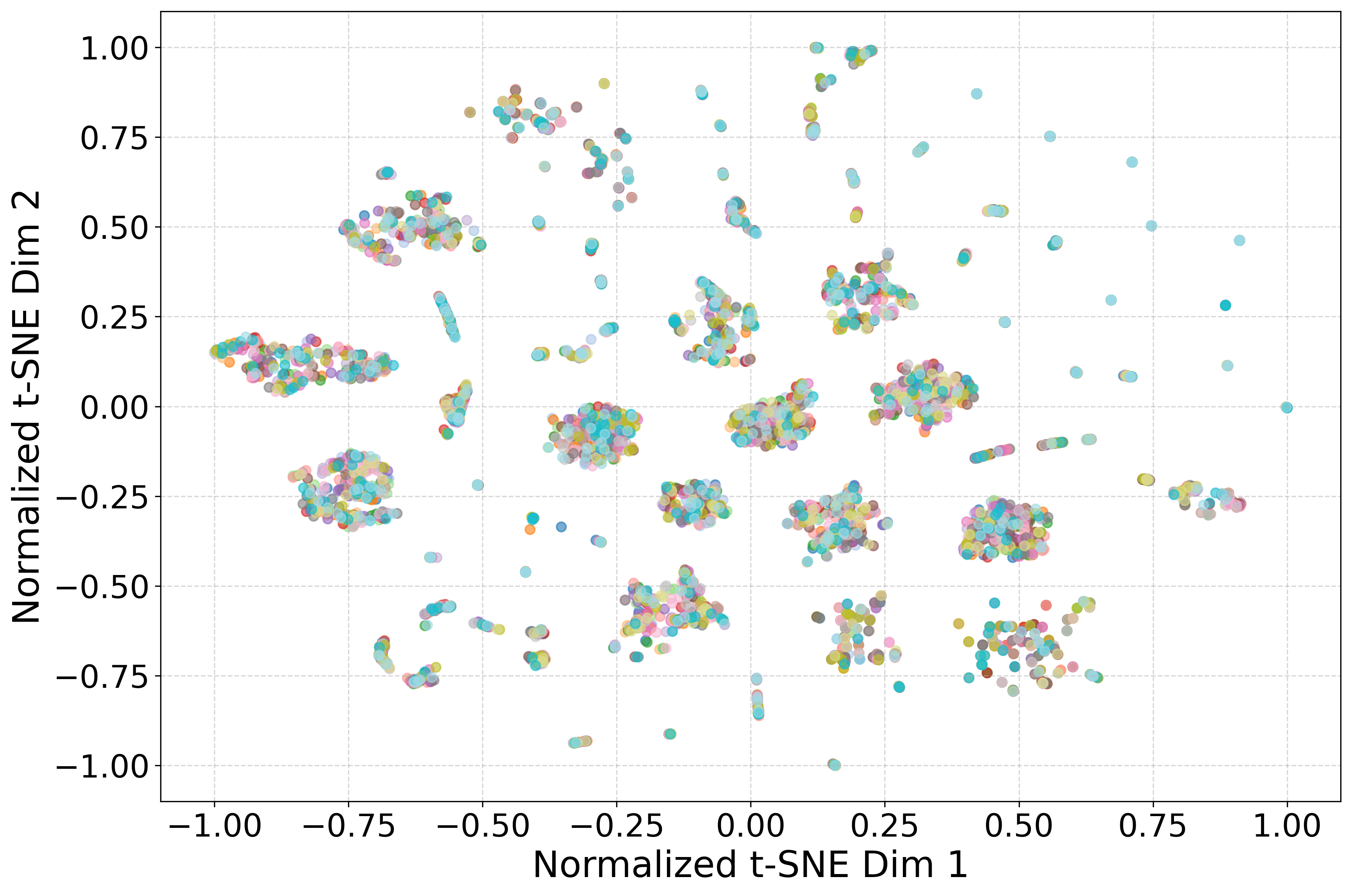}
    \caption{BBH embeddings.}
    \label{fig:bbh_embedding}
  \end{subfigure}
  \hfill
  \begin{subfigure}[b]{0.33\textwidth}
    \centering
    \includegraphics[width=0.99\textwidth]{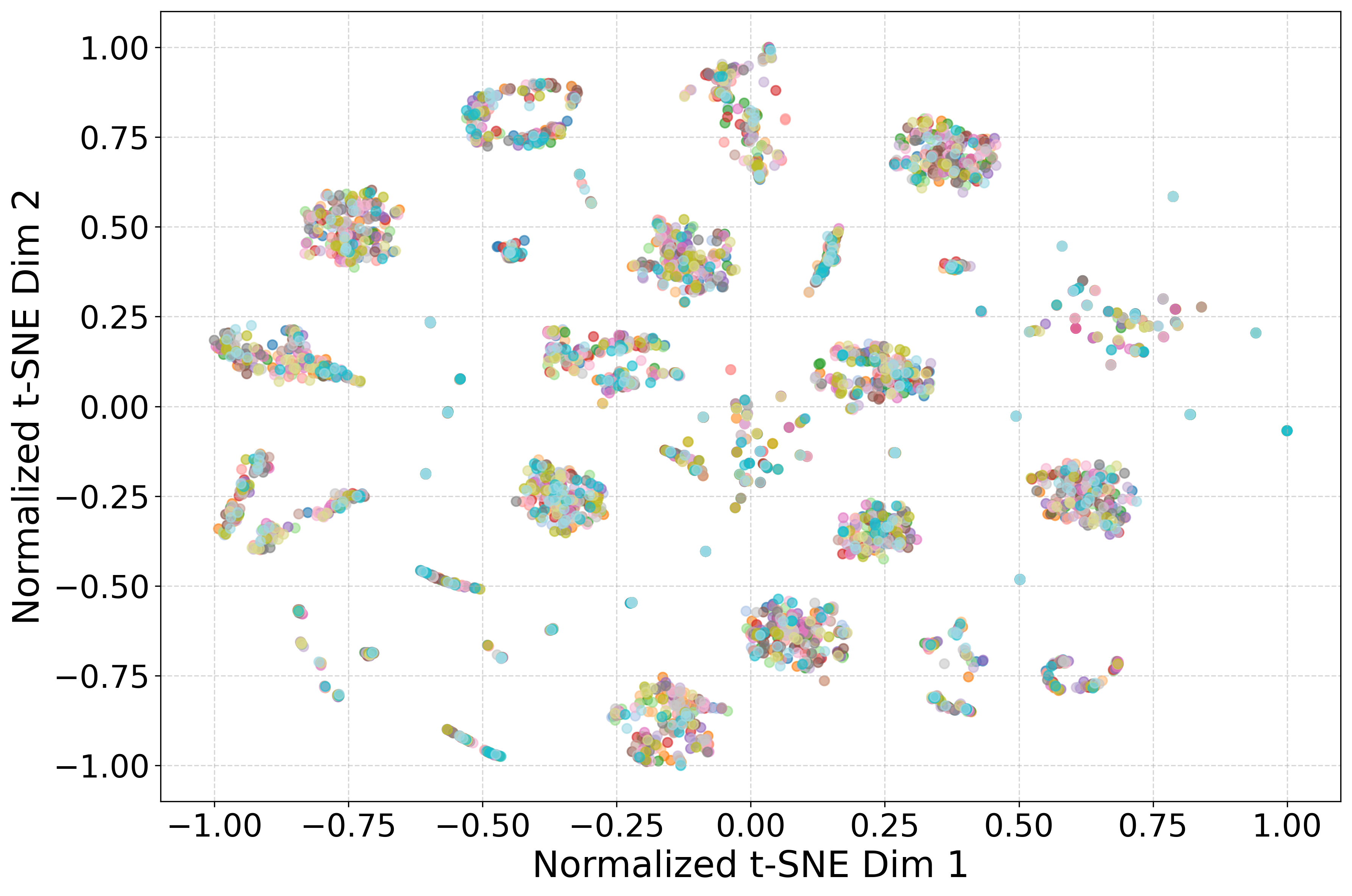}
    \caption{BBEH embeddings.}
    \label{fig:bbeh_embedding}
  \end{subfigure}
  \caption{T-SNE visualization of query embeddings for \textsc{General365} and comparative benchmarks. 
  (a) \textsc{General365} (1460 instances) shows a highly dispersed and uniform distribution, highlighting the semantic independence achieved through our seed problems. (b) BBH (6511 instances) and (c) BBEH (4520 instances) exhibit significant "local collapse" with isolated, high-density clusters.}
  \label{fig:embedding}
\end{figure}

\paragraph{Qualitative Analysis of Semantic Space Coverage.} 
To qualitatively assess the semantic breadth and distributional characteristics of \textsc{General365}, we project high-dimensional query embeddings into a 2D manifold. Specifically, we leverage the \texttt{text-embedding-ada-002} model to encode queries into 1536-dimensional vectors, followed by t-SNE (\texttt{Perplexity}=30) for dimensionality reduction. For visual clarity, coordinates are normalized to the range $[-1, 1]$. 
As illustrated in Figure ~\ref{fig:embedding}, \textsc{General365} exhibits a substantially more uniform and expansive distribution across the semantic space compared to existing benchmarks. In contrast, BBH and BBEH manifest significant "local collapse," forming isolated, high-density clusters. This phenomenon suggests that while these benchmarks contain numerous samples, they are often confined to a narrow set of linguistic templates or logical skeletons. Conversely, the spatial dispersion of \textsc{General365} underscores the efficacy of our high-quality seed strategy, where human-curated seed instances ensure semantic independence and minimize redundancy at the source.

\paragraph{Quantitative Analysis of LLM-based Reasoning Similarity Scoring.} 
To rigorously quantify the logical redundancy within the benchmarks, we propose a pairwise similarity scoring method. Unlike qualitative visualization, this approach deconstructs the underlying reasoning primitives and logic evolution of task instances. 
Specifically, for each problem $T_A$ in \textsc{General365}, we retrieve its nearest neighbor $T_B$ in the embedding space\footnote{Embeddings are generated using the \texttt{text-embedding-ada-002} model.} to identify the most semantically similar candidate. 
Then we employ Gemini-3-Pro\footnote{Gemini-3-Pro demonstrates exceptional performance across diverse reasoning tasks, exhibiting significant potential in evaluating the semantic diversity and logical redundancy of problems in the benchmark.} as an expert evaluator to conduct a deep comparison of the triplets (Problem, CoT, Final Answer) for each pair. 
The similarity is quantified on a scale of $0$ to $5$, where $0$ denotes total logical independence and $5$ indicates essentially identical reasoning skeletons, the scoring template is shown in Appendix ~\ref{sec:prompts}. 
As summarized in Figure ~\ref{fig:diversity_scored}, \textsc{General365} demonstrates a markedly left-skewed distribution, with a mean similarity score of only $\mu=2.16$. Approximately $68.2\%$ of its instances fall within the $1$-$2$ score range, suggesting that even semantically related queries in \textsc{General365} maintain distinct logical paths. 
In contrast, BBH ($\mu=4.71$) and BBEH ($\mu=4.80$) exhibit extreme right-skewness, with $77.8\%$ and $83.5\%$ of their samples respectively receiving the maximum score of $5$. 
These findings empirically verify that while existing benchmarks often expand via template-based perturbations—resulting in high structural homogeneity, \textsc{General365} preserves a diverse spectrum of independent reasoning patterns, effectively minimizing cognitive redundancy.

\begin{figure}[t]
    \centering
    \includegraphics[width=0.7\textwidth]{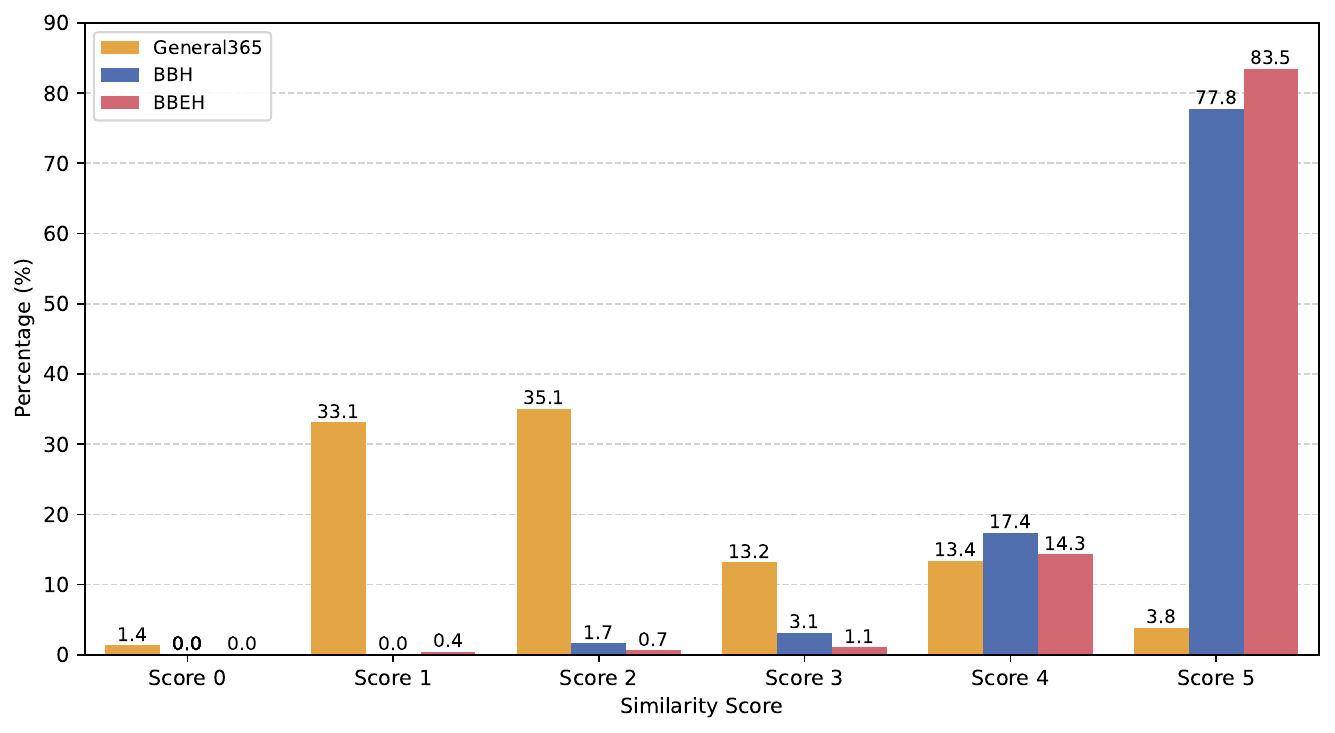} 
    \caption{The distribution of reasoning similarity scores across benchmarks shows a clear right-skewed distribution for BBH and BBEH, while \textsc{General365} exhibits a left-skewed distribution.}
    \label{fig:diversity_scored}
\end{figure}

\subsubsection{Validation of Difficulty}
\begin{wrapfigure}{r}{0.60\textwidth}
    \centering
    \includegraphics[width=0.60\textwidth]{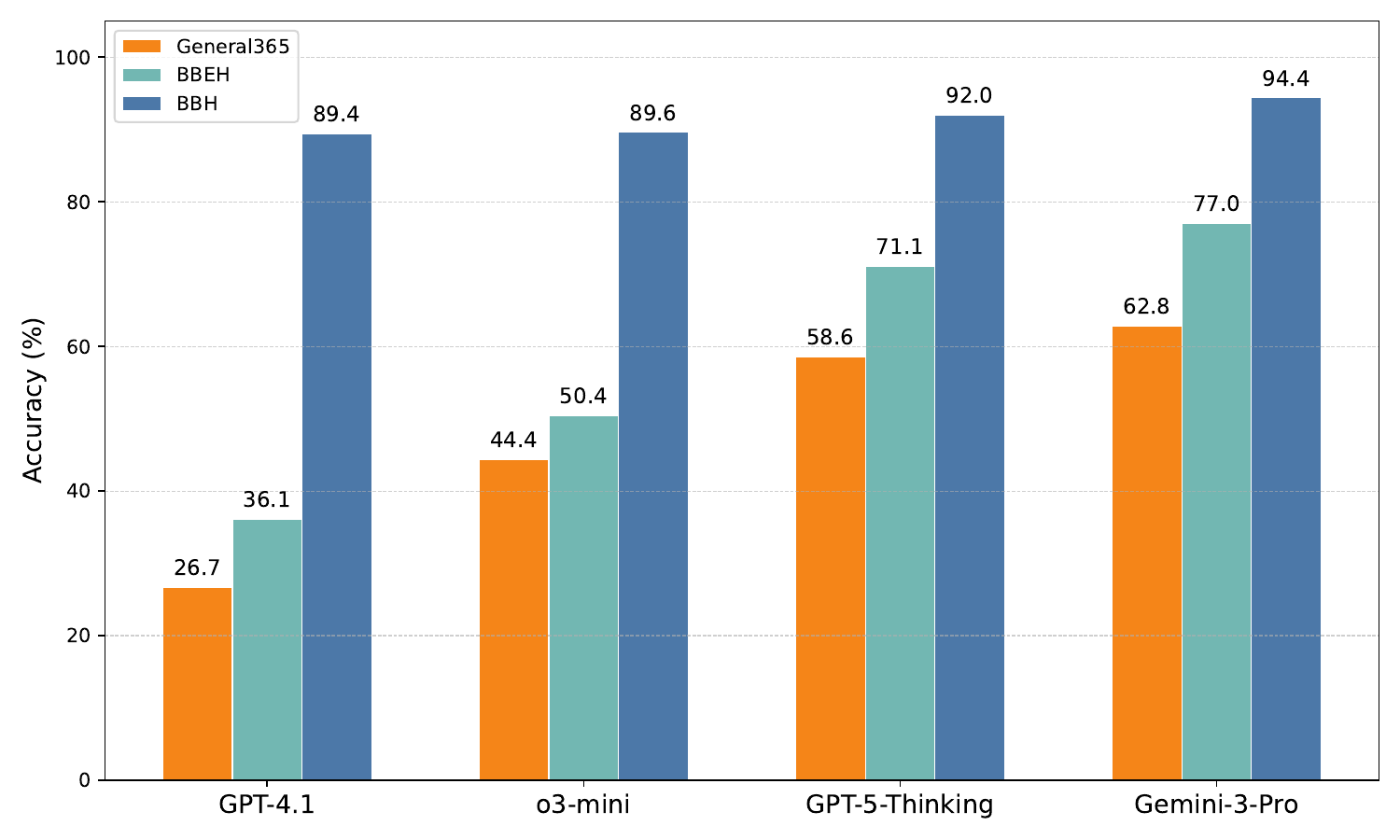} 
    \caption{Comparative performance across benchmarks, highlighting the significantly higher difficulty floor of \textsc{General365}.}
    \label{fig:compare_accuracy}
\end{wrapfigure}

\paragraph{\textsc{General365} as a High-Difficulty Benchmark for General Reasoning.}
As shown in Figure ~\ref{fig:compare_accuracy}, the overall accuracy on \textsc{General365} remains substantially lower than that on BBH and BBEH across all evaluated models, indicating a significantly higher difficulty. 
Furthermore, compared to BBH and BBEH, \textsc{General365} exhibits a much wider performance gap between models of varying capability levels, effectively distinguishing the reasoning boundaries of current LLMs in real-world general tasks where existing benchmarks have become almost saturated.

\begin{figure}[t]
    \centering
    \includegraphics[width=0.99\textwidth]{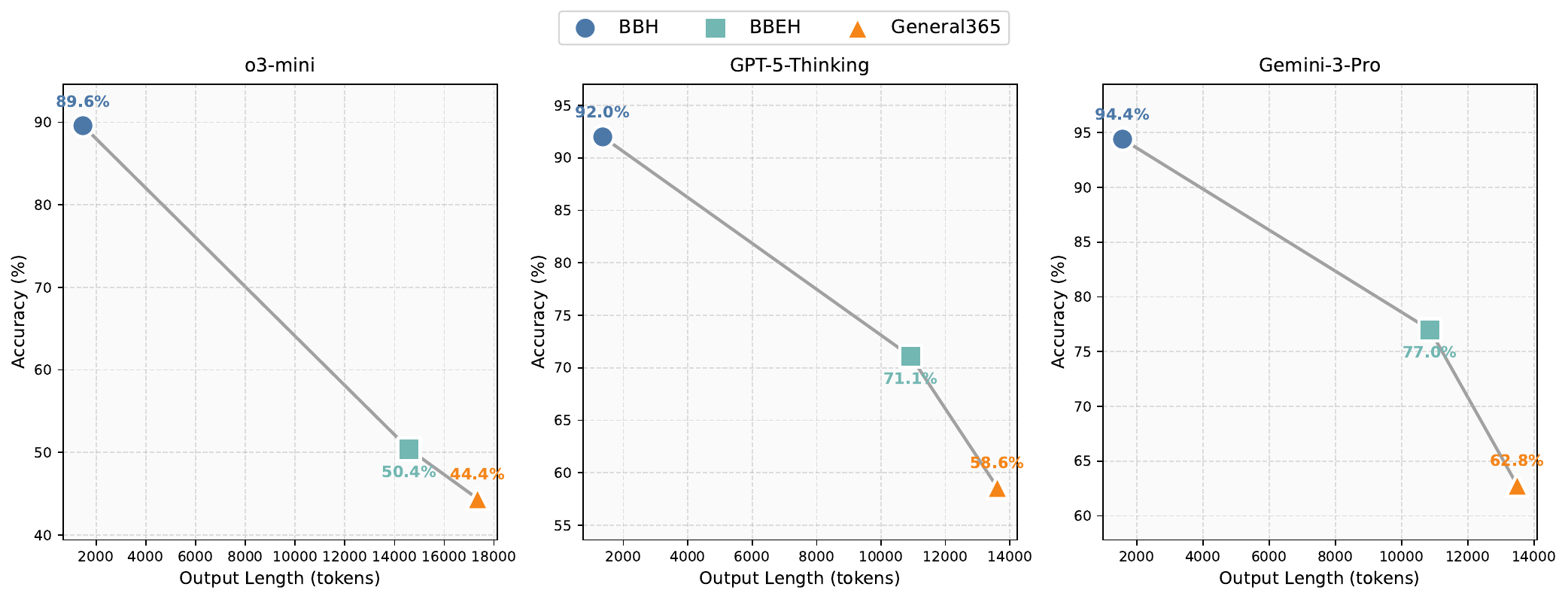}
    \caption{The relationship between accuracy and average output length varies significantly across benchmarks: the fact that models yield substantially longer outputs on \textsc{General365} despite lower accuracy underscores the benchmark's rigorous demand for deep reasoning.}
    \label{fig:output_length}
\end{figure}

\paragraph{The high difficulty is further evidenced by the models' average output length.} 
Figure \ref{fig:output_length} illustrates a clear inverse correlation between model performance and average response length during evaluation. 
Specifically, across top-tier models such as GPT-5-Thinking and Gemini-3-Pro, benchmarks with higher accuracy (e.g., BBH) correspond to significantly more concise outputs, often requiring fewer than 2k tokens for successful completion. 
In stark contrast, \textsc{General365} consistently demands the longest reasoning length over 13k tokens for GPT-5-Thinking and Gemini-3-Pro, and exceeding 18k for o3-mini (despite yielding lower accuracy scores). 
This divergence suggests that \textsc{General365} tasks necessitate more elaborate, multi-step logical derivations and finer-grained thinking paths. 
These findings demonstrate that the increased token usage in \textsc{General365} is not merely a sign of verbosity but a direct consequence of the taxing cognitive load and intricate reasoning trajectories required to navigate its high-difficulty problems.

\section{Related Work}
Evaluating LLMs on specific domains such as mathematics, physics, programming or other subjects has been a critical aspect for assessing advancements in reasoning capabilities. In mathematics, AMO-bench~\citep{Amobench} and FrontierMath~\citep{frontiermath} challenge models with competition-level problems that push the boundaries of expert-level quantitative reasoning. Within the physical sciences, Abench-physics~\citep{Abench-physics} and Physunibench~\citep{Physunibench} assess the models' capacity to navigate complex physical laws and multi-step scientific derivations. 
Similarly, in the programming domain, SWE-bench~\citep{Swe-bench} and its successor SWE-bench-pro~\citep{Swe-bench-pro} provide rigorous frameworks for evaluating the resolution of real-world software engineering issues. 
Furthermore, as part of the effort to evaluate graduate-level expertise, GPQA~\citep{gpqa} and Super-GPQA~\citep{supergpqa} present benchmarks comprising challenging multiple-choice problems authored by domain experts. 
Concurrently, HLE~\citep{phan2025humanity} constructs a multi-subject, closed-ended academic benchmark to serve as a last exam for LLMs, challenging them with problems that demand both advanced expertise and rigorous logical inference.
However, we argue that the performance on these benchmarks often stems from the models' ability to memorize intricate patterns within specialized training corpora rather than a mastery of fundamental logical rules. This heavy reliance on domain-specific knowledge obscures the assessment of a model's true inferential ability, as the underlying reasoning process is frequently conflated with knowledge recall.

Diverging from the emphasis on domain-specific knowledge, a burgeoning line of work has shifted focus toward the generalization of reasoning capabilities within broader, everyday domains. 
These works aim to evaluate LLMs' proficiency in executing complex reasoning tasks within a general domain, where the cognitive load stems from intricate logical structures rather than the retrieval of domain-specific knowledge. 
For instance, BBH~\citep{bbh} distills a challenging subset of tasks from the original BIG-bench~\citep{bigbench} collection, specifically targeting reasoning dimensions where LLMs initially failed to surpass human-level performance; BBEH~\citep{bbeh} further amplifies the cognitive demand by introducing tasks with significantly higher structural complexity, designed to expose the reasoning boundary of even the SOTA models; 
ARC-AGI~\citep{arc} presents a collection of unique visual reasoning tasks that require models to infer underlying procedural rules from only a few input-output examples, emphasizing the ability to generalize to novel patterns without relying on prior linguistic or domain-specific knowledge; 
KOR-Bench~\citep{korbench} focuses on knowledge-oriented reasoning, evaluating LLMs' capacity to execute complex logical operations and synthesize information across diverse, knowledge-intensive domains. 
While BBH and ARC-AGI offer task diversity, their average difficulty has become insufficient due to the rapid advancement of LLMs, leading to a performance saturation. 
Conversely, although BBEH and KOR-Bench present substantial challenges, they are often constrained by narrow task coverage and template-based instance construction. 
Such structural rigidity risks performance inflation, where models may leverage spurious reasoning shortcuts or hacks rather than demonstrating reasoning depth.

\section{Conclusion}
In this paper, we introduce \textsc{General365}, a diverse and challenging benchmark designed to evaluate the general reasoning capabilities of LLMs in broad, real-world domains. Through a meticulous construction pipeline (incorporating human-curated seed problems and model-based augmentation with rigorous quality audits), we establish a dataset characterized by high structural diversity and substantial cognitive challenge. 
Extensive evaluations across 26 leading LLMs reveal that contemporary models still struggle significantly with general reasoning, with the state-of-the-art model achieving only a 62.8\% accuracy. 
Furthermore, our fine-grained analysis identifies "Semantic Interference" and "Optimal Strategy" as the primary performance bottlenecks for current architectures. We expect that \textsc{General365} will serve as a catalyst for future research, inspiring the community to push LLM reasoning beyond domain-specific expertise and toward the development of robust, generalized reasoning frameworks.

\section*{Acknowledgments}
We thank Dongyu Ru, Zijian Zhang, Yuhuai Wei, Zeyang Yu, Jiaxing Liu, Jiaming Wang, Yiyang Li, Linsen Guo, Cunguang Wang, and Zhimin Lin for their insightful suggestions regarding the construction and analysis of \textsc{General365}.
We are grateful to Wei Wang, Wenjie Shi, Jiaqi Zhang, Xiangyu Xi, Xiangzhou Huang, Rongxiang Weng, Jingang Wang, Xin Chen, Jianhuan Li, and Keheng Wang for the valuable discussions and insights on model performance.
We also appreciate the engineering support provided by Yunke Zhao and Dengchang Zhao, and open-source assistance from Qi Li and Peng Wang.

\bibliographystyle{unsrtnat}
\bibliography{references}

\clearpage

\appendix
\section{Representative Examples of Challenge Categories}\label{appendix:examples}
This section presents representative examples for each of the eight challenge categories identified in Section~\ref{sec:challenge}. These cases are curated to illustrate the specific reasoning challenges and inherent complexities that characterize each category.

\begin{exmp}{Complex Constraints Problem}{text_answer}
{}\textbf{Question:} A, B, C, D, E, and F six people went to a coffee shop to catch up. They sat on both sides of a square table, with three people on each side. Each person ordered a different type of coffee. It is known that:  \par
~~~~~1. B was sitting next to C.  \par
~~~~~2. C was sitting opposite the person adjacent to E.  \par
~~~~~3. D was sitting opposite A, and D ordered cappuccino.  \par
~~~~~4. Those who ordered iced espresso are sitting opposite E.  \par
~~~~~5. Neither D nor F ordered mocha.  \par
~~~~~6. F did not order cold brew coffee.  \par
~~~~~7. The person who ordered cold brew coffee was sitting opposite D.  \par
~~~~~8. The person sitting next to B ordered Irish coffee.  \par
Among them, there was one person who ordered espresso. How many possible seating arrangements were there for these six people at that time? \par
\textbf{Answer:} \boxed{8}
\end{exmp}


\begin{exmp}{Branching \& Enumeration Problem}{text_answer}
{}\textbf{Question:} There are 8 boxes, labeled A through H. A total of 8 cartons of milk are distributed among them. We know the following: Exactly 5 of the 8 boxes contain milk (meaning 3 boxes are empty, containing 0 cartons). Any box that contains milk has at least one carton. \par
~~~~~The total number of cartons in boxes A, B, C, and D is 5. \par
~~~~~The total number of cartons in boxes D, E, and F is 4. \par
~~~~~The total number of cartons in boxes C and D is 2. \par
Based on the information above, among all milk distribution schemes, how many involve Box A containing milk? \par
\textbf{Answer:} \boxed{16}
\end{exmp}

\begin{exmp}{Spatial \& Temporal Reasoning Problem}{text_answer}
{}\textbf{Question:} Xiaokang accidentally entered a maze filled with thick smoke. He couldn't see anything clearly and couldn't distinguish directions. He could only follow the arrows on the ground to move from one square to another.
Xiaokang remembered his route as follows: \par
~~~~~1. Move forward one step, then move one step to the upper right, and turn left 45 degrees;\par
~~~~~2. Move forward one step, then move one step to the upper left, and turn right 135 degrees;\par
~~~~~3. Move forward three steps, then move one step to the lower right, and turn left 45 degrees;\par
~~~~~4. Move forward three steps, then move two steps to the right, and turn right 90 degrees.\par
~~~~~$\rightarrow$ $\searrow$ $\swarrow$ $\leftarrow$ $\downarrow$ \par
~~~~~$\uparrow$ $\leftarrow$ $\rightarrow$ $\nearrow$ $\downarrow$ \par
~~~~~$\leftarrow$ $\leftarrow$ $\leftarrow$ $\leftarrow$ $\downarrow$ \par
~~~~~$\swarrow$ $\nwarrow$ $\leftarrow$ $\leftarrow$ $\nwarrow$ \par
~~~~~$\swarrow$ $\searrow$ $\rightarrow$ $\rightarrow$ $\nwarrow$ \par
Where is Xiaokang now in the diagram? \par
\textbf{Answer:} The starting point
\end{exmp}

\begin{exmp}{Recursive \& Backtracking Problem}{text_answer}
{}\textbf{Question:} There are 9 tunnels in a row, and one of them hides an enemy wounded soldier. The wounded soldier will move to an adjacent tunnel after a certain period of time; during this period, you only have enough time to check one tunnel. What is the minimum number of checks required to ensure you catch the wounded soldier? \par
\textbf{Answer:} \boxed{14}
\end{exmp}

\begin{exmp}{Semantic Interference Problem}{text_answer}
{}\textbf{Question:} In another parallel universe, affected by Event A in 2130, the concepts of leap years and common years are reversed. Due to the C effect caused by Event A, compared to the Gregorian calendar era in our spacetime, the calendar system of a native tribe in this parallel universe has now become the same as ours. Originally, their calendar system had the opposite number of days in months compared to our Gregorian calendar except for February (That is to say, months that originally had 31 days would become 30 days, and so on). \par
~~~~~In 2070, this parallel universe was affected by a special policy B, under which all people decided to increase the number of days in odd-numbered months by one day and decrease the number of days in even-numbered months by one day, for all calendar systems at any time before and after this moment. \par
~~~~~For two individuals from the aforementioned native tribe in this parallel universe, assume that the number of days in March 2000 is a and the number of days in February 2105 is b, what is the value of 10a + b? \par
\textbf{Answer:} \boxed{327}
\end{exmp}

\begin{exmp}{Implicit Information Reasoning Problem}{text_answer}
{}\textbf{Question:} Xiaoyi came across an interesting problem. She tried ranking by population and by total area, but both attempts failed. She thought, perhaps both of these failed approaches still have their merits? Do you know what the result is?
Please answer the number represented by the question mark. \par
~~~~~1. Australia \^{} Brazil = 7776 \par
~~~~~2. India - Canada = 5 \par
~~~~~3. China + United States = 7 \par
~~~~~4. Argentina * Kazakhstan = ? \par
\textbf{Answer:} \boxed{72}
\end{exmp}

\begin{exmp}{Optimal Strategy Problem}{Optimal Strategy}
{}\textbf{Question:} A fire truck receives an emergency call and heads to the fire scene. The starting point is 10 km away from the destination. The road speed limit is 60 km/h, and the fire truck’s theoretical maximum average speed is 80 km/h. It is expected that there will be 3 intersections with standard red-yellow-green traffic lights along the way. Each traffic light cycle has a red light for 58 seconds, a green light for 58 seconds, and a yellow light for 4 seconds. The light the fire truck encounters when it reaches each intersection is completely random, but it is guaranteed that it will encounter at least one full red light. \par
~~~~~Ignoring traffic congestion and any issues of legality, what is the shortest possible time, in minutes, for the fire truck to reach the destination? \par
\textbf{Answer:} \boxed{7.5}
\end{exmp}

\begin{exmp}{Probability \& Uncertainty Problem}{select_answer}
{}\textbf{Question:} After Dong Da's sudden death, Yu Bai attended his friend's memorial service with great sorrow. All the friends were present that day, except for Haitang and Xiao Liu. During the memorial service, a gunshot was suddenly heard. Yu Bai and Er Dan quickly rushed to the scene of the shooting and saw Xiao Liu lying crookedly on the ground. Yu Bai quickly secured the scene and stopped others from entering, exclaiming sadly, "We've lost a friend. The murderer is likely hiding in the crowd. Please be careful." Di Di said, "Let's call the police." Suddenly, Detective Da Ji stepped out of the crowd and said: "I already know who killed Xiao Liu. It must be Haitang, because she wanted to avenge Dong Da!" Yu Bai glanced at him in surprise and said, "I already know who the murderer is. Let's call the police quickly." \par
~~~~~Who is the most likely suspect?   \par
~~~~~A. Haitang   \par
~~~~~B. Er Dan   \par
~~~~~C. Yu Bai   \par
~~~~~D. Detective \par
\textbf{Answer:} D
\end{exmp}

\section{Prompt Templates}\label{sec:prompts}

\paragraph{Pairwise similarity scoring template.}
We employ the LLM-based pairwise similarity scoring using Gemini-3-Pro as the scoring model, and use the following prompt to quantify the logical redundancy within \textsc{General365} in Section~\ref{sec:validation}.

\begin{exmp}{Prompt Template For Pairwise Similarity Scoring}{query_prompt}
    \textbf{\#\#\# Role:} \\
    You are a distinguished expert in Cognitive Science and General AI Reasoning Logic. Your specialty lies in measuring the essential similarity between tasks by deconstructing their \#underlying\_logic, \#knowledge\_systems, and \#cognitive\_challenge\_types, rather than relying on superficial semantic descriptions.

    \textbf{\#\#\# Objective:} \\
    Perform a deep deconstruction and comparison of [Task A] and [Task B] (including the Problem, Chain-of-Thought (CoT) reasoning, and Final Result). Output a similarity score on a scale of 0-5. \\
    CRITICAL REQUIREMENT: Focus your analysis on the logic evolution trajectory within the CoT, rather than the surface-level wording of the problem.

    \textbf{\#\#\# Evaluation Dimensions:} \\
    \# 1. Core\_Knowledge\_Graph: Overlap in theorems, formulas, or facts. \\
    \# 2. Reasoning\_Paradigm: Chaining, branching, or network logic (Inductive/Deductive, etc.). \\
    \# 3. Source\_of\_Challenge: Difficulty sources like hidden conditions, reverse thinking, or logic depth. \\
    \# 4. Operational\_Primitives: Atomic actions like symbolic transformation, case analysis, or recursion.

    \textbf{\#\#\# Scoring Standard (0-5 Scalar):} \\
    0 - Irrelevant: Zero overlap in logic/knowledge. \\
    1 - Weakly Related: Macro-level connection (e.g., both "math") but CoT paths are entirely different. \\
    2 - Superficially Similar: Same topic but fundamentally different reasoning depths. \\
    3 - Logically Overlapping: Significant overlap in reasoning logic or operators, though domains differ. \\
    4 - Highly Structurally Similar: Identical "reasoning skeleton" and challenge sources. \\
    5 - Essentially Identical: Variants of each other; logic paths and operators are seamlessly transferable.

    \textbf{\#\#\# Output Format (Strict Adherence Required):} \\
    Final Similarity Score: [A scalar number between 0 and 5]

    \textbf{\#\#\# Task Inputs:} \\{}
    [Task A]: \{task1\} \\{}
    [Task B]: \{task2\}
\end{exmp}

\paragraph{Query prompt template.}
In order to guide LLMs in generating answers in a rule-readable format, we use the following prompt template (e.g., Example~\ref{exmp:query_prompt_num}) to guide model generation. The prompt templates for problems with select and text answers are also shown below (e.g., Example~\ref{exmp:query_prompt_select} and Example~\ref{exmp:query_prompt_text}).

\begin{exmp}{Template For Problems With Numerical Answer}{query_prompt_num}
    ...
    
    Output your final answer at the end of your reply using the following format:
    
    \textbf{\#\#\# The final answer is:
    \$\textbackslash boxed\{\textless your answer\textgreater\}\$}
    
    Example:
    
    \textbf{\#\#\# The final answer is:
    \$\textbackslash boxed\{123\}\$ or \$\textbackslash boxed\{(1, 2]\}\$}

    The final answer should be given as precisely as possible (using LaTeX symbols such as \textbackslash sqrt, \textbackslash frac, \textbackslash pi, etc.).
    If the final answer involves a decimal approximation, it must be accurate to at least FLOAT\_ROUND decimal places.

\end{exmp}

\begin{exmp}{Template For Problems With Select Answer}{query_prompt_select}
    ...
    
    Output your final answer at the end of your reply using the following format:
    
    \textbf{\#\#\# The final answer is:
    \$ \textless your answer\textgreater ~\$}

    Select one or more appropriate options as your final answer based on the question above.

\end{exmp}

\begin{exmp}{Template For Problems With Text Answer}{query_prompt_text}
    ...
    
    Output your final answer at the end of your reply using the following format:
    
    \textbf{\#\#\# The final answer is:
    \$ \textless your answer\textgreater ~\$}

\end{exmp}

\paragraph{Final answer grading prompt template.}
We employ the model-based grading using GPT-4.1 as the grading model, and use the following grading prompt to verify the equivalence between the LLM output and the reference answer.

\begin{exmp}{Grading Prompt Template For Final Answer}{grading_prompt}
For the following problem, we have the reference answer and the student's answer.

Determine whether the student's answer is equivalent to the reference answer.

If equivalent, output "Correct".

If not equivalent, output "Incorrect".\\

\#\#\# Problem

...\\

\#\#\# Reference Answer

...\\

\#\#\# Student Answer

...\\

Now, please provide your judgment.

Please strictly follow the format below to summarize your conclusion at the end of your judgment:

\#\#\# Conclusion: Correct/Incorrect

If the answer involves a decimal approximation, it must be accurate to at least four decimal places.
\end{exmp}

\section{Performance on the Public Subset of \textsc{General365}}\label{sec:white-box}
In order to facilitate community research and ensure reproducibility, we have released a publicly accessible subset of \textsc{General365}. 
Specifically, this public split is constructed by randomly sampling 180 seed problems along with their extended variants, yielding a total of 720 evaluation instances. 
This section details the performance of various models on this open-access subset. 
The detailed results are summarized in Table ~\ref{tab:general365_open-access_subset}.

\begin{table*}[t]
\centering
\caption{Performance of various LLMs on the complete \textsc{General365} benchmark and its publicly accessible subset. The closely aligned scores demonstrate that the public subset reliably reflects models' overall capabilities on the full benchmark. }
\label{tab:general365_open-access_subset}
\small
\begin{tabular}{ccc}
\toprule
\textbf{Model Name} & \textbf{Acc (\%)} & \textbf{Acc (Public Subset) (\%)} \\ \midrule
Gemini-3-Pro & 62.8 & 61.3 \\
Gemini-3-Flash & 60.8 & 59.7 \\
Gemini-2.5-Pro & 48.7 & 50.0 \\
Gemini-2.5-Flash & 39.6 & 38.2 \\ 
GPT-5.1-Thinking & 58.2 & 57.1 \\
GPT-5-Thinking & 58.6 & 56.0 \\
o4-mini & 51.0 & 51.4 \\
o3-mini & 44.4 & 44.4 \\
Claude-Sonnet-4.5 & 48.6 & 46.8 \\
Grok-4.1-Fast-Reasoning & 53.1 & 51.5 \\ 
Qwen3.5-397B-A17B-Thinking & 57.7 & 57.9 \\
Qwen3-Max-Thinking & 57.2 & 57.9 \\
Qwen3-235B-Thinking-2507 & 47.9 & 46.0 \\
Qwen3-Max-Instruct & 48.5 & 48.1 \\
DeepSeek-V3.2-Speciale & 57.5 & 56.4 \\
DeepSeek-V3.2-Thinking & 54.9 & 55.4 \\
DeepSeek-V3.1-Thinking & 48.1 & 47.1 \\
DeepSeek-V3.2-Chat & 37.6 & 37.2 \\
Kimi-K2.5-Thinking & 56.4 & 57.1 \\
Kimi-K2-Thinking & 53.0 & 53.1 \\ 
GLM-5-Thinking & 59.9 & 59.2 \\
GLM-4.7-Thinking & 57.4 & 55.8 \\
GLM-4.6-Thinking & 52.9 & 52.6 \\
LongCat-Flash-Thinking-2601 & 50.8 & 50.0 \\
LongCat-Flash-2512 & 41.3 & 40.3 \\
\bottomrule
\end{tabular}
\end{table*}

\section{Robustness to Sampling Variance}\label{sec:sample-variance}
To address the high diversity and inherent difficulty of \textsc{General365}, we employ a relatively high sampling temperature during inference, encouraging models to explore diverse reasoning pathways. 
While this strategy is beneficial for tackling complex tasks, it inevitably introduces stochasticity that can lead to noticeable variability in the final evaluation outcomes. 
In this section, we investigate the impact of this sampling variance by conducting four independent evaluation runs for a set of representative models. 
We measure the performance fluctuations using the maximum deviation ($\Delta$ = max-min). 
As detailed in Table \ref{tab:stability_analysis}, the deviations remain consistently under 3\% across different architectures, confirming that our evaluation metrics are reliable and highly reproducible despite the high-temperature setting.

\begin{table}[h]
\centering
\caption{Stability analysis of representative models on the \textsc{General365} across four independent evaluation runs. We report the accuracy (\%) for individual repetitions and the maximum deviation ($\Delta$ = max-min) between runs.}
\label{tab:stability_analysis}
\small
\begin{tabular}{cccccc}
\toprule
\textbf{Model Name} & \textbf{Rep 1} & \textbf{Rep 2} & \textbf{Rep 3} & \textbf{Rep 4 (Acc)} & \boldmath$\Delta$  \\ \midrule
DeepSeek-V3.2-Thinking & 53.4 & 54.9 & 55.1 & 54.9 & 1.7 \\
Kimi-K2-Thinking & 51.5 & 53.2 & 53.0 & 53.0 & 1.7 \\
GLM-4.7-Thinking & 57.0 & 55.4 & 57.4 & 57.4 & 2.0 \\
LongCat-Flash-Thinking-2601 & 50.7 & 50.9 & 51.6 & 50.8 & 0.9 \\
LongCat-Flash-2512 & 40.4 & 43.0 & 41.0 & 41.7 & 2.6 \\
\bottomrule
\end{tabular}
\end{table}

\end{document}